\documentclass[letterpaper]{article} % DO NOT CHANGE THIS
\usepackage{aaai2026}  % DO NOT CHANGE THIS
\usepackage{times}  % DO NOT CHANGE THIS
\usepackage{helvet}  % DO NOT CHANGE THIS
\usepackage{courier}  % DO NOT CHANGE THIS
\usepackage[hyphens]{url}  % DO NOT CHANGE THIS
\usepackage{graphicx} % DO NOT CHANGE THIS
\usepackage{natbib}  % DO NOT CHANGE THIS AND DO NOT ADD ANY OPTIONS TO IT
\usepackage{caption} % DO NOT CHANGE THIS AND DO NOT ADD ANY OPTIONS TO IT
\usepackage{algorithm}
\usepackage{algorithmic}
\usepackage{amsmath,amssymb,amsthm}
\usepackage{multirow}
\usepackage{subcaption}
\usepackage{booktabs}
\usepackage{newfloat}
\usepackage{listings}
\DeclareCaptionStyle{ruled}{labelfont=normalfont,labelsep=colon,strut=off} % DO NOT CHANGE THIS
\floatstyle{ruled}
\newfloat{listing}{tb}{lst}{}
\floatname{listing}{Listing}
\theoremstyle{plain}
\newtheorem{theorem}{Theorem}[]

\theoremstyle{definition}

\theoremstyle{remark}

\title{Beyond Sharpness: A Flatness Decomposition Framework for Efficient Continual Learning}
\author{
    Yanan Chen\textsuperscript{\rm 1,\rm 2}, 
    Tieliang Gong\textsuperscript{\rm 1,\rm 2}\thanks{Corresponding author.}, 
    Yunjiao Zhang\textsuperscript{\rm 4}, 
    Wen Wen\textsuperscript{\rm 1,\rm 3}
}
\affiliations{
    \textsuperscript{\rm 1}School of Computer Science and Technology, Xi’an Jiaotong University\\
    \textsuperscript{\rm 2}Ministry of Education Key Laboratory of Intelligent Networks and Network Security, Xi’an Jiaotong University\\
    \textsuperscript{\rm 3}Shaanxi Province Key Laboratory of Big Data Knowledge Engineering, Xi’an Jiaotong University\\
    \textsuperscript{\rm 4}China Telecom\\
    \{ynchen894, adidasgtl\}@gmail.com
}

\begin{document}

\maketitle

\begin{abstract}
Continual Learning (CL) aims to enable models to sequentially learn multiple tasks without forgetting previous knowledge. Recent studies have shown that optimizing towards flatter loss minima can improve model generalization.
However, existing sharpness-aware methods for CL suffer from two key limitations: (1) they treat sharpness regularization as a unified signal without distinguishing the contributions of its components. and (2) they introduce substantial computational overhead that impedes practical deployment. 
To address these challenges, we propose FLAD, a novel optimization framework that decomposes sharpness-aware perturbations into gradient-aligned and stochastic-noise components, and show that retaining only the noise component promotes generalization.  We further introduce a lightweight scheduling scheme that enables FLAD to maintain significant performance gains even under constrained training time. FLAD can be seamlessly integrated into various CL paradigms and consistently outperforms standard and sharpness-aware optimizers in diverse experimental settings, demonstrating its effectiveness and practicality in CL.
\end{abstract}

% Uncomment the following to link to your code, datasets, an extended version or similar.
% You must keep this block between (not within) the abstract and the main body of the paper.
% \begin{links}
%     \link{Code}{https://aaai.org/example/code}
%     \link{Datasets}{https://aaai.org/example/datasets}
%     \link{Extended version}{https://aaai.org/example/extended-version}
% \end{links}

\section{Introduction}

Continual Learning (CL) aims to enable models to learn from a non-stationary stream of data or tasks while retaining previously acquired knowledge \cite{hadsell_embracing_2020,de_lange_continual_2022}. This setting poses a fundamental challenge known as catastrophic forgetting, a phenomenon where the learning of new tasks causes abrupt degradation in performance on previously learned ones \cite{parisi_continual_2019}.
Recent effort has been devoted to addressing this challenge by designing diverse paradigms including replay-based methods \cite{rebuffi_icarl_2017,lin_pcr_2023}, regularization-based methods \cite{lin_pcr_2023,sun_decoupling_2023} and architecture-based methods \cite{hu_dense_2023,zhou_model_2023}. 

While these paradigms have proven effective, they often impose substantial storage requirements for raw data or model parameters, raising scalability concerns as task complexity increases \cite{zhou_model_2023,wang_comprehensive_2024}. An alternative research direction focuses on enhancing the intrinsic generalization capacity of models to address these limitations. This has motivated exploration of sharpness-aware minimization methods, which seek flat minima that are widely believed to yield superior generalization and robustness under distributional shifts \cite{keskar_large-batch_2017,zhang_gradient_2023}.
Notable examples include Sharpness-Aware Minimization (SAM) \cite{foret_sharpness-aware_2021}, which leverages zeroth-order sharpness to seek minima that are locally robust to parameter perturbations, and Gradient norm Aware Minimization (GAM) \cite{zhang_gradient_2023}, which exploits first-order sharpness to improve feature reuse. These approaches have shown considerable promise in transfer and few-shot learning, and have recently been adapted to CL with specialized designs \cite{tran_tung_sharpness_2023,chen_sharpness-aware_2024,bian_make_2024}.

Despite their empirical success, existing flatness-based methods in CL suffer from two fundamental limitations that constrain their practical applicability. First, these approaches adopt an overly simplistic treatment of sharpness regularization, conceptualizing it as a monolithic objective rather than recognizing the heterogeneous nature of loss landscape curvature. Specifically, they fail to disentangle the distinct geometric properties and learning implications of different curvature components, particularly the critical distinction between gradient-aligned and gradient-orthogonal directions in parameter space. Second, they require substantial architectural modifications to standard optimization pipelines, introducing prohibitive computational overhead that severely limits their scalability and real-world deployment. These approaches typically require either double-gradient computations through higher-order differentiation or multiple forward-backward passes per optimization step, which significantly increases the training time.

In this paper, we propose FLAD, a novel Flatness Decomposition framework for continual learning that addresses the aforementioned limitations. Specifically, FLAD decomposes perturbation directions in sharpness-aware minimization by isolating and penalizing only the stochastic-noise component, thereby preventing counterproductive penalties on directions essential for optimization. Our comprehensive empirical analysis demonstrates that this targeted approach enables models to escape sharp valleys more effectively, converging to flatter minima with enhanced generalization properties. Furthermore, recognizing the computational inefficiencies of existing methods, particularly under time-constrained training scenarios, we introduce a lightweight optimization scheme that maximizes learning efficacy without incurring additional computational overhead.

In summary, our main contributions are as follows:% 
\begin{itemize}
    \item We propose a CL framework that leverages a decomposed flatness signal to escape sharp minima and preserve past knowledge 
    with negligible computational overhead.
    \item We provide a novel interpretation of sharpness-aware optimization by decomposing perturbation directions into gradient-aligned and stochastic-noise components, and demonstrate the critical role of noise-aligned perturbations in improving generalization for CL.
    \item Extensive experimental results demonstrate that our framework outperforms state-of-the-art competitors across multiple CL benchmarks. We further show that even partial application of our strategy yields significant gains, enabling substantial reductions in training time without sacrificing effectiveness.
    
\end{itemize} 

\section{Related Works}

\paragraph{Continual learning.} CL methods are broadly categorized into three groups \cite{wang_comprehensive_2024,van_de_ven_three_2022,parisi_continual_2019}: replay-based, regularization-based, and architecture-based methods. Replay-based methods mitigate forgetting by extending data space with selected exemplars from old tasks~\cite{rebuffi_icarl_2017,rolnick_experience_2019,sun_regularizing_2023}. Regularization-based methods introduce explicit constraints into the loss function to balance performance between new and old tasks \cite{lin_pcr_2023,saha_gradient_2021,sun_decoupling_2023}, which typically act on the parameter space or the gradients to preserve past knowledge during optimization. Architecture-based methods allocate task-specific sub-networks or expand the model, thereby minimizing interference across tasks \cite{lu_revisiting_2024,zhou_model_2023}. Alternatively, recent efforts  explicitly modify the training dynamics to preserve generalization across tasks \cite{wu_meta_2024,lee_learning_2024}. One direction is to modify the gradients of different tasks to overcome forgetting \cite{chaudhry_efficient_2019,lopez-paz_gradient_2017}. Another emerging line of research investigates CL from the perspective of the loss landscape, aiming to improve generalization by steering training towards flatter minima \cite{shi_overcoming_2021,deng_flattening_2021}. Several works have empirically shown the benefits of SAM in mitigating catastrophic forgetting and preserving performance on previous tasks \cite{mehta_empirical_2023,chen_sharpness-aware_2024}, including DFGP \cite{yang_data_2023}, FS-DGPM \cite{deng_flattening_2021}, and SAM-CL series \cite{tran_tung_sharpness_2023}. Recent work C-Flat combines zeroth- and first-order sharpness to achieve globally flatter solutions and improve CL performance \cite{bian_make_2024}. While promising, such sharpness-aware methods remain underexplored in CL. 
% , demonstrates that it helps
\paragraph{Sharpness-aware minimization.} 
Sharpness-aware minimization was originally proposed to improve generalization by guiding optimization toward flat minima in standard supervised learning 
\cite{foret_sharpness-aware_2021,he_asymmetric_2019}.
This has motivated increasing interest in understanding the theoretical foundations of its generalization behavior, especially those of SAM, which is based on zeroth order sharpness. These effort include PAC-Bayes generalization analyses \cite{foret_sharpness-aware_2021}, studies of implicit bias and optimization dynamics \cite{andriushchenko_towards_2022,chen_sharpness-aware_2024,chen_why_2023}, and recent investigations dissecting the core components responsible for empirical effectiveness in SAM \cite{li_friendly_2024}. These empirical and theoretical developments motivate our work, propose a principled and efficient framework for CL that leverages the core mechanism of sharpness-aware optimization to improve generalization.

\section{Method}

In this section, we propose a framework for continual learning that explicitly incorporates flatness-promoting perturbations. By leveraging both zeroth- and first-order sharpness information, the method aims to mitigate forgetting while enabling effective knowledge transfer across tasks.

\paragraph{Loss landscape sharpness.} Let \(\mathcal{D}\) denote the training distribution on \(\mathcal{X}\times\mathcal{Y}\) and \(S = \{(x_i, y_i)\}^n_{i=1}\) denote the training dataset with \(n\) data-points drawn independently from \(\mathcal{D}\). Let a family of models parameterized by \(w\in\mathcal{W}\in\mathbb{R}^d\) and a per-data-point loss function \(\ell(\cdot):\mathcal{W}\times\mathcal{X}\times\mathcal{Y}\mapsto \mathbb{R}^+\). The empirical training loss is defined as
\begin{equation}
    L(w)=\frac{1}{n}\sum_{i=1}\ell_i(w).
    \label{eq1}
\end{equation}
Zeroth-order sharpness is the worst-case loss within a defined neighborhood, defined over training set \(S\) as 
\cite{foret_sharpness-aware_2021}:
\begin{equation}
    R^{0}_{\rho}(w)=\underset{\| \delta\|_2<\rho}{\text{max}}L(w+\delta)-L(w),
    \label{eq2}
\end{equation}
where \(\rho>0\) denotes the neighborhood radius.

First-order sharpness is a measurement of the maximal neighborhood gradient norm, reflecting landscape curvature and defined over training set \(S\) as \cite{zhang_gradient_2023}: 
\begin{equation}
    R^{1}_{\rho}(w)=\underset{\| \delta\|_2<\rho}{\text{max}}\nabla L(w+\delta),
    \label{eq3}
\end{equation}
where \(\nabla L (w)\) is the gradient of \(L(w)\) with respect to \(w\).

\paragraph{Optimization with sharpness.} To efficiently optimize the zeroth-order flatness in Eq. \ref{eq2}, it is approximated via first-order expansion and compute the adversarial perturbation \(\delta\) as follows \cite{foret_sharpness-aware_2021}:
\begin{equation}
    \delta=\rho \cdot \frac{\nabla L (w)}{\|\nabla L (w)\|_2}.
    \label{eq4}
\end{equation}
Subsequently, one can compute the gradient at the perturbed point \(w+\delta\), and then use the updating step of the base optimizer such as SGD to update
\begin{equation}
    w_{t+1}=w_t-\lambda\nabla L (w)|_{w_t+\delta_t}.
    \label{eq5}
\end{equation}

The zeroth-order sharpness minimization addresses the loss landscape curvature through the gradient norm \(\|\nabla L_B(w)\|\) in Eq. \ref{eq4}, which serves as a proxy for local sharpness. To enhance generalization, we aim to remove the shared, dominated structure from this direction and retain only its batch-specific stochastic variation \cite{li_friendly_2024}. 
Denote the batch gradient by \(\nabla L_B(w)\) as \(\hat{g}_B\), and the full-batch gradient \(\nabla L(w)\) by \(g\), we have
\begin{equation}
    \hat{g}_B=\text{Proj}_{g}(\hat{g}_B)+\text{Proj}^{\top}_{g}(\hat{g}_B),
    \label{eq6}
\end{equation}
where \(\text{Proj}_g(\hat{g}_B)\) is the projection of \(\hat{g}_B\) onto \(g\), \(\text{Proj}^{\top}_g(\hat{g}_B)\) is the orthogonal component. In practice, we approximate the orthogonal component as 
\begin{equation}
    \text{Proj}_{g}^{\top}\hat{g}_B=\hat{g}_B-\sigma m_t,
    \label{eq7}
\end{equation}
where \(\sigma=\cos(g,\hat{g}_B)\) denotes the cosine similarity between \(g\) and \(\hat{g}_B\). To improve training efficiency, \(\sigma\) is fixed as a constant during optimization.

Note that computing \(g\) over the full dataset in each iteration is computationally prohibitive. We therefore maintain an exponential moving average (EMA) of the per-step gradient norm to approximate the full-batch gradient:
\begin{equation}
    m_t=\lambda_0 m_{t-1}+(1-\lambda_0)\hat{g}_{B_t},
    \label{eq8}
\end{equation}
where \(\lambda_0\in (0,1)\) is a hyper-parameter. 
It is proved that EMA provides a reliable approximation of the full gradient \cite{li_friendly_2024}. We can use \(\text{Proj}_{g}^{\top}\hat{g}_B\) Eq. \ref{eq7} to substitute \(g\) in Eq. \ref{eq4} to get \(\delta\) then update in Eq. \ref{eq5}.

Similarly, the minimization of the first-order sharpness in Eq.\ref{eq3} is approximated and compute the adversarial perturbation, follow by the updating step
\begin{align}
    \delta&=\rho\frac{\nabla\|\nabla L (w)\|}{\|\nabla\|\nabla L (w)\|_2\|_2}, \nonumber \\
    w_{t+1}&=w_t-\lambda\nabla\|\nabla L (w)\|_2|_{w_t+\delta_t},
    \label{eq9}
\end{align}
where \(\nabla\|\nabla L (w)\|\) denotes the gradient of \(\|\nabla L (w)\|\) with respect to \(w\). Note that
\begin{equation}
    \nabla\|\nabla L (w)\|=\nabla^2 L(w)\cdot\frac{\nabla L(w)}{\|\nabla L(w)\|}.
    \label{eq10}
\end{equation}

While the zeroth-order method targets the norm of the gradient, the first-order variant considers the sharpness of the gradient itself, that is, the curvature of the function \(\|L(w)\|\). 
As before, we use \(\hat{g}_B\) and \(g\) to denote \(\nabla L_B(w)\) and \(\nabla L(w)\). To extract meaningful stochastic directions, we decompose \(\nabla|\hat{g}_B\|\) into components aligned with and orthogonal to the EMA-estimated full-batch direction:
\begin{align}
    \nabla\|\hat{g}_B\|=\text{Proj}_{\nabla\|g\|}\nabla&\|\hat{g}_B\| + \text{Proj}^{\top}_{\nabla\|g\|}\nabla\|\hat{g}_B\|, \nonumber \\
    \text{Proj}_{\nabla\|g\|}^{\top}\nabla\|\hat{g}_B\| &= \nabla\|\hat{g}_B\|- \sigma\nabla\|g\|,
    \label{eq11}
\end{align}
where \(\sigma\) is a constant during training, \(n_t\) is the estimation of the gradient sharpness direction:
\begin{equation}
     n_t=\lambda_1 n_{t-1}+(1-\lambda_1)\nabla\|\hat{g}_{B_t}\|,
    \label{eq12}
\end{equation}
where \(\lambda_1\in(0,1)\) is a hyperparameter. 

We unify both zeroth- and first-order sharpness minimization in a single optimization framework, where the perturbation directions in each are respectively decomposed into stochastic-noise components. 
The optimization is detailed in Algorithm \ref{algorithm1}. By using Hessian-vector product in Eq. \ref{eq10}, we significantly reduce both time and memory complexity, the overall strategy requires 2 forward and 4 backward passes per iteration. The framework introduces minimal additional computational overhead and can be seamlessly integrated
into standard optimizers.

\begin{algorithm}[tb]
\caption{FLAD Algorithm}
\label{algorithm1}
\textbf{Input}: Training phase \(T\), training data \(S^T\), model \(f^{T-1}\) with parameter \(w^{T-1}\) from last phase if \(T > 1\), batchsize \(b\), oracle loss function \(L\), learning rate \(\eta > 0\), neighborhood size \(\rho\), trade-off coefficient \(\gamma\), hyper-parameter \(\lambda_0,\lambda_1,\sigma\), small constant \(c\).\\
\textbf{Output}: Model trained at the current time \(T\).

\begin{algorithmic}[1] %[1] enables line numbers
\STATE \textbf{Initialize:} \textbf{if} \(T=1\): Randomly Initialize parameter \\\(w^{T=1}\), \(t\gets0\), \(m_{-1}=n_{-1}=0\)
\WHILE{\(w^{T}\) not converge,}
\STATE Sample batch \(B^{T}\) of b from \(S^{T}\)
\STATE Compute batch’s loss gradient \(\hat{g}_{B_t}=\nabla L_{B^T}(w)\)
\STATE Compute \(m_t\) in Eq. \ref{eq8}
\STATE Compute perturbation \(\delta_0=\rho\frac{\hat{g}_{B_t}-\sigma m_t}{\| \hat{g}_{B_t}-\sigma m_t\|+c}\)
\STATE Approximate \(g_0=\nabla L_{B^T}(w+\delta_0)\)
\STATE Compute Hessian vector product in Eq. \ref{eq10}
\STATE Compute \(n_t\) in Eq. \ref{eq12}
\STATE Compute \(\delta_1=\rho\frac{\nabla\|\hat{g}_{B_t}\|-\sigma n_t}{\| \nabla\|\hat{g}_{B_t}\|-\sigma n_t\|+c}\)
\STATE Approximate \\ \(\qquad\) \(g_1=\nabla^2 L_{B^T}(w+\delta_1)\cdot\frac{\nabla L_{B^T}(w+\delta_1)}{\|\nabla L_{B^T}(w+\delta_1)\|+c}\)
\STATE Update: model parameter: \(w=w-\eta(g_0+\gamma g_1)\)
\ENDWHILE
\STATE \textbf{return} \(w^T\) and \(f^T\)
\end{algorithmic}
\end{algorithm}

\begin{figure*}[htbp]
	\centering
	\begin{subfigure}{0.31\linewidth}
		\centering
		\includegraphics[width=1\linewidth]{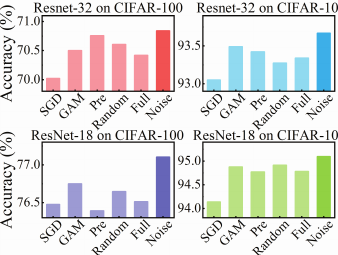}
		\caption{Effect of perturbation}
		\label{2a}%文中引用该图片代号
	\end{subfigure}
	\centering
	\begin{subfigure}{0.42\linewidth}
		\centering
		\includegraphics[width=1\linewidth]{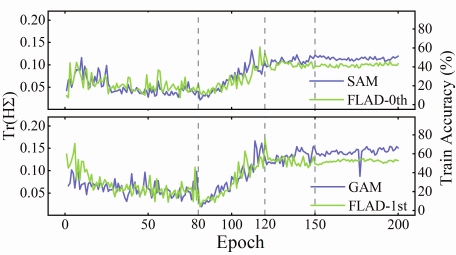}
		\caption{Evolution of \(\text{Tr}(\mathbf{H\Sigma})\)}
		\label{2b}%文中引用该图片代号
	\end{subfigure}
    \centering
	\begin{subfigure}{0.26\linewidth}
		\centering
		\includegraphics[width=1\linewidth]{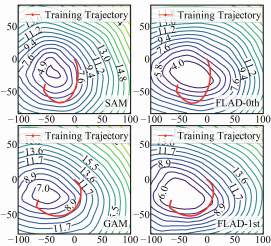}
		\caption{2D Loss Landscape}
		\label{2c}%文中引用该图片代号
	\end{subfigure}
	\caption{Empirical analysis of perturbation components in zeroth- and first-order sharpness minimization.
(a) Average accuracy across four training settings using different perturbation directions. 
(b) \(\text{Tr}(\mathbf{H\Sigma})\) during training for stochastic-noise and original variants. 
(c) 2D loss landscape visualizations around final model parameters and training trajectory.}
	\label{fig2}
\end{figure*}

\paragraph{Continual learning.}

We now integrate our noise-aware optimizer into a general continual learning framework, focusing on the class-incremental learning (CIL) setting, which is arguably the most challenging CL scenario, where the model must continuously acquire knowledge from sequentially presented, class-agnostic data without task identity at inference.

Our approach is broadly compatible with the three major paradigms in CIL: replay-based, regularization-based, and expansion-based methods. Replay-based and regularization-based methods can directly benefit from our optimizer by plugging the original task loss in the curvature-regularized objective:
\begin{equation}
    L(w^T)=L^{R_\rho^0}(w^T)+\gamma\cdot L^{R_\rho^1}(w^T),
    \label{eq14}
\end{equation}
where the zeroth- and first-order terms encourage flat minima aligned with generalization. These methods typically operate on data streams that includes exemplars, and our optimizer fits naturally into this training protocol using FLAD Algorithm. Expansion-based methods extend the model by allocating task-specific modules for new classes. In this case, the same objective in Eq. \ref{eq14} can be applied to the added components during the learning phase, while shared components remain frozen. Final inference is performed after a post-processing step, such as classifier calibration or gating.

This formulation demonstrates the flexibility of our method in adapting to various CL strategies, while preserving its benefits centered on flatness-induced generalization.

\subsection{Empirical Analysis}

To validate the effectiveness behind our unified framework, we conduct an empirical investigation into the underlying mechanisms of zeroth- and first-order sharpness minimization by 1) identifying the effective components in the adversarial perturbation that contributes to improved generalization, and 2) quantitatively analyze the gradient-aligned and stochastic-noise components of the perturbations, providing deeper insights into the role of stochasticity in optimization.

\begin{table*}[tb]
    \centering
    \setlength{\tabcolsep}{2.2mm}{
    \small
    \begin{tabular}{c c c|c c|c c|c c|c}
    \toprule
        \multirow{3}{*}{Methods} & \multicolumn{2}{c}{CIFAR-10} & \multicolumn{4}{c}{CIFAR-100}                           & \multicolumn{2}{c}{Tiny-ImageNet} & Average \\  \cmidrule(lr){2-3} \cmidrule(lr){4-7} \cmidrule(lr){8-9} %\Xcline{2-9}{0.4pt}
        ~ & \multicolumn{2}{c}{N=5}     & \multicolumn{2}{c}{N=10}  & \multicolumn{2}{c}{N=5}   & \multicolumn{2}{c}{N=8}  & Return        \\  \cmidrule(lr){2-3} \cmidrule(lr){4-5} \cmidrule(lr){6-7} \cmidrule(lr){8-9} % \Xcline{2-9}{0.4pt}
        ~ & AAA & \multicolumn{1}{c}{Acc} & AAA & \multicolumn{1}{c}{Acc} & AAA & \multicolumn{1}{c}{Acc} & AAA & \multicolumn{1}{c}{Acc} & AAA  \\ \midrule
        
        \multicolumn{1}{c|}{Replay} 
        & 61.84 {\tiny\(\pm\)6.66} & 41.68 {\tiny\(\pm\)3.61} & 49.89 {\tiny\(\pm\)1.04} & 29.07 {\tiny\(\pm\)1.30} & 53.94 {\tiny\(\pm\)1.72} & 33.62 {\tiny\(\pm\)0.36} & 39.51 {\tiny\(\pm\)0.47} & 19.09 {\tiny\(\pm\)1.17} \\
        \multicolumn{1}{c|}{w/SAM} & 60.91 {\tiny\(\pm\)5.42} & 42.57 {\tiny\(\pm\)4.32} & \underline{51.16} {\tiny\(\pm\)1.13} & \underline{31.31} {\tiny\(\pm\)2.12} & \underline{54.91} {\tiny\(\pm\)1.59} & \textbf{34.97} {\tiny\(\pm\)0.49} & 41.19 {\tiny\(\pm\)0.27} & 21.22 {\tiny\(\pm\)0.60} \\
        \multicolumn{1}{c|}{w/GAM} & \underline{61.92} {\tiny\(\pm\)5.52} & \underline{43.07} {\tiny\(\pm\)6.82} & 50.26 {\tiny\(\pm\)0.93} & 30.30 {\tiny\(\pm\)1.39} & 54.49 {\tiny\(\pm\)1.76} & \textbf{34.97} {\tiny\(\pm\)1.39} & 40.10 {\tiny\(\pm\)0.51} & 20.23 {\tiny\(\pm\)0.39} \\
        \multicolumn{1}{c|}{w/C-Flat} & 60.46 {\tiny\(\pm\)6.62} & 42.32 {\tiny\(\pm\)2.85} & 51.15 {\tiny\(\pm\)1.27} & 31.07 {\tiny\(\pm\)1.86} & 54.85 {\tiny\(\pm\)1.65} & \underline{34.40} {\tiny\(\pm\)0.38} & \underline{41.95} {\tiny\(\pm\)0.10} & \underline{22.29} {\tiny\(\pm\)0.20} \\
        \multicolumn{1}{c|}{w/FLAD} & \textbf{62.60} {\tiny\(\pm\)4.85} & \textbf{43.13} {\tiny\(\pm\)4.32} & \textbf{51.81} {\tiny\(\pm\)0.95} & \textbf{31.74} {\tiny\(\pm\)1.69} & \textbf{55.19} {\tiny\(\pm\)1.59} & \underline{34.40} {\tiny\(\pm\)0.10} & \textbf{42.41} {\tiny\(\pm\)0.18} & \textbf{22.91} {\tiny\(\pm\)0.75} & +2.18\%\\

        \midrule
        \multicolumn{1}{c|}{iCaRL} 
        & 66.54 {\tiny\(\pm\)4.42} & 50.63 {\tiny\(\pm\)3.17} & 51.11 {\tiny\(\pm\)1.23} & 30.05 {\tiny\(\pm\)1.83} & 56.84 {\tiny\(\pm\)1.72} & 35.91 {\tiny\(\pm\)0.76} & 41.12 {\tiny\(\pm\)0.49} & 20.38 {\tiny\(\pm\)0.53} \\
        \multicolumn{1}{c|}{w/SAM} & 66.85 {\tiny\(\pm\)4.57} & 50.93 {\tiny\(\pm\)4.86} & \underline{51.45} {\tiny\(\pm\)0.96} & \underline{30.14} {\tiny\(\pm\)1.98} & \underline{57.33} {\tiny\(\pm\)1.87} & \underline{36.95} {\tiny\(\pm\)0.62} & 42.35 {\tiny\(\pm\)0.27} & 21.83 {\tiny\(\pm\)0.40} \\
        \multicolumn{1}{c|}{w/GAM} & 66.46 {\tiny\(\pm\)4.82} & 49.86 {\tiny\(\pm\)3.94} & 51.19 {\tiny\(\pm\)1.06} & 30.05 {\tiny\(\pm\)2.20} & 56.84 {\tiny\(\pm\)1.52} & 36.07 {\tiny\(\pm\)0.59} & 42.00 {\tiny\(\pm\)0.48} & 21.23 {\tiny\(\pm\)0.68} \\
        \multicolumn{1}{c|}{w/C-Flat} & \underline{67.89} {\tiny\(\pm\)4.10} & \underline{51.62} {\tiny\(\pm\)6.38} & 51.09 {\tiny\(\pm\)1.21} & 30.05 {\tiny\(\pm\)1.83} & 57.21 {\tiny\(\pm\)1.90} & 36.75 {\tiny\(\pm\)0.10} & \underline{42.81} {\tiny\(\pm\)0.41} & \underline{22.17} {\tiny\(\pm\)0.35} \\
        \multicolumn{1}{c|}{w/FLAD} & \textbf{68.32} {\tiny\(\pm\)4.54} & \textbf{52.53} {\tiny\(\pm\)6.02} & \textbf{51.75} {\tiny\(\pm\)1.21} & \textbf{30.36} {\tiny\(\pm\)2.06} & \textbf{57.59} {\tiny\(\pm\)1.60} & \textbf{37.13} {\tiny\(\pm\)0.97} & \textbf{43.29} {\tiny\(\pm\)0.57} & \textbf{23.04} {\tiny\(\pm\)0.86} & +1.24\%\\

        \midrule
        \multicolumn{1}{c|}{WA}  
        & 72.03 {\tiny\(\pm\)3.58} & 61.95 {\tiny\(\pm\)4.33} & 62.56 {\tiny\(\pm\)1.26} & 47.29 {\tiny\(\pm\)0.25} & 66.89 {\tiny\(\pm\)2.00} & 53.59 {\tiny\(\pm\)0.21} & 51.92 {\tiny\(\pm\)0.38} & 36.40 {\tiny\(\pm\)0.23} \\
        \multicolumn{1}{c|}{w/SAM} & \underline{72.27} {\tiny\(\pm\)3.94} & 61.26 {\tiny\(\pm\)5.77} & 63.04 {\tiny\(\pm\)1.45} & 47.35 {\tiny\(\pm\)0.98} & 67.36 {\tiny\(\pm\)2.17} & 54.55 {\tiny\(\pm\)0.73} & 52.39 {\tiny\(\pm\)0.07} & \underline{37.81} {\tiny\(\pm\)0.33} \\
        \multicolumn{1}{c|}{w/GAM} & 69.50 {\tiny\(\pm\)4.70} & \underline{62.18} {\tiny\(\pm\)2.78} & \underline{63.13} {\tiny\(\pm\)1.16} & \underline{47.79} {\tiny\(\pm\)0.69} & \underline{67.38} {\tiny\(\pm\)2.10} & 54.51 {\tiny\(\pm\)0.43} & 50.95 {\tiny\(\pm\)1.15} & 36.56 {\tiny\(\pm\)0.46} \\
        \multicolumn{1}{c|}{w/C-Flat} & 71.59 {\tiny\(\pm\)1.95} & 61.45 {\tiny\(\pm\)4.26} & 62.24 {\tiny\(\pm\)1.57} & 47.04 {\tiny\(\pm\)1.02} & 67.24 {\tiny\(\pm\)1.98} & \underline{54.66} {\tiny\(\pm\)0.34} & \underline{52.71} {\tiny\(\pm\)0.84} & 35.61 {\tiny\(\pm\)0.08} \\
        \multicolumn{1}{c|}{w/FLAD} & \textbf{72.43} {\tiny\(\pm\)1.80} & \textbf{62.46} {\tiny\(\pm\)4.26} & \textbf{63.73} {\tiny\(\pm\)1.19} & \textbf{48.35} {\tiny\(\pm\)0.74} & \textbf{67.63} {\tiny\(\pm\)2.02} & \textbf{54.69} {\tiny\(\pm\)0.47} & \textbf{53.53} {\tiny\(\pm\)0.18} & \textbf{39.00} {\tiny\(\pm\)0.15} & +1.90\%\\
        \midrule
        
        \multicolumn{1}{c|}{FOSTER}  
        & 65.15 {\tiny\(\pm\)3.86} & 56.43 {\tiny\(\pm\)4.55} & 52.71 {\tiny\(\pm\)1.25} & \textbf{39.71} {\tiny\(\pm\)1.68} & 56.32 {\tiny\(\pm\)1.74} & 37.91 {\tiny\(\pm\)0.55} & 48.32 {\tiny\(\pm\)0.66} & 38.10 {\tiny\(\pm\)0.55} \\
        \multicolumn{1}{c|}{w/SAM} & \underline{74.80} {\tiny\(\pm\)3.26} & 68.52 {\tiny\(\pm\)2.78} & 52.40 {\tiny\(\pm\)1.89} & 37.81 {\tiny\(\pm\)1.80} & \underline{56.47} {\tiny\(\pm\)2.30} & 37.92 {\tiny\(\pm\)0.14} & 48.49 {\tiny\(\pm\)1.15} & \underline{38.42} {\tiny\(\pm\)1.79} \\
        \multicolumn{1}{c|}{w/GAM} & 63.86 {\tiny\(\pm\)5.31} & 54.04 {\tiny\(\pm\)3.30} & \underline{52.93} {\tiny\(\pm\)1.09} & \underline{39.41} {\tiny\(\pm\)1.95} & 56.19 {\tiny\(\pm\)1.75} & \underline{38.09} {\tiny\(\pm\)0.52} & 48.49 {\tiny\(\pm\)1.03} & 38.27 {\tiny\(\pm\)1.00} \\
        \multicolumn{1}{c|}{w/C-Flat} & 73.68 {\tiny\(\pm\)3.62} & \underline{68.81} {\tiny\(\pm\)4.65} & 52.05 {\tiny\(\pm\)1.84} & 38.71 {\tiny\(\pm\)1.85} & 56.35 {\tiny\(\pm\)2.00} & 37.81 {\tiny\(\pm\)0.66} & \underline{48.56} {\tiny\(\pm\)0.96} & 38.25 {\tiny\(\pm\)0.55} \\
        \multicolumn{1}{c|}{w/FLAD} & \textbf{75.08} {\tiny\(\pm\)2.17} & \textbf{68.98} {\tiny\(\pm\)4.41} & \textbf{53.00} {\tiny\(\pm\)2.01} & 39.13 {\tiny\(\pm\)1.80} & \textbf{56.51} {\tiny\(\pm\)2.01} & \textbf{38.82} {\tiny\(\pm\)0.85} & \textbf{48.66} {\tiny\(\pm\)0.88} & \textbf{38.54} {\tiny\(\pm\)1.15} & +1.41\%\\

        \midrule
        \multicolumn{1}{c|}{MEMO} 
        & \underline{67.86} {\tiny\(\pm\)6.63} & 57.80 {\tiny\(\pm\)4.55} & 64.38 {\tiny\(\pm\)2.83} & 53.27 {\tiny\(\pm\)4.16} & 68.05 {\tiny\(\pm\)1.94} & \textbf{61.58} {\tiny\(\pm\)8.40} & 56.53 {\tiny\(\pm\)0.55} & 44.29 {\tiny\(\pm\)0.18} \\
        \multicolumn{1}{c|}{w/SAM} & 65.99 {\tiny\(\pm\)6.51} & 54.96 {\tiny\(\pm\)2.96} & \underline{65.41} {\tiny\(\pm\)3.00} & 53.67 {\tiny\(\pm\)0.67} & 69.10 {\tiny\(\pm\)1.69} & 59.40 {\tiny\(\pm\)0.25} & \underline{57.53} {\tiny\(\pm\)0.66} & \underline{44.64} {\tiny\(\pm\)0.27} \\
        \multicolumn{1}{c|}{w/GAM} & 66.92 {\tiny\(\pm\)6.69} & 55.34 {\tiny\(\pm\)5.56} & 64.60 {\tiny\(\pm\)2.72} & 53.22 {\tiny\(\pm\)0.82} & 68.36 {\tiny\(\pm\)1.85} & 57.78 {\tiny\(\pm\)1.23} & 57.22 {\tiny\(\pm\)0.84} & 44.43 {\tiny\(\pm\)0.27} \\
        \multicolumn{1}{c|}{w/C-Flat} & 66.29 {\tiny\(\pm\)7.34} & \underline{58.04} {\tiny\(\pm\)4.80} & 64.96 {\tiny\(\pm\)2.63} & \textbf{54.07} {\tiny\(\pm\)0.74} & \underline{68.95} {\tiny\(\pm\)1.41} & 59.70 {\tiny\(\pm\)0.40} & 57.41 {\tiny\(\pm\)1.32} & 44.38 {\tiny\(\pm\)0.23} \\
        \multicolumn{1}{c|}{w/FLAD} & \textbf{68.51} {\tiny\(\pm\)6.07} & \textbf{61.40} {\tiny\(\pm\)4.92} & \textbf{65.47} {\tiny\(\pm\)2.75} & \underline{53.92} {\tiny\(\pm\)0.48} & \textbf{69.30} {\tiny\(\pm\)1.50} & \underline{60.09} {\tiny\(\pm\)0.06} & \textbf{57.91} {\tiny\(\pm\)0.95} & \textbf{45.06} {\tiny\(\pm\)0.12} & +1.83\%\\

        \midrule
        \multicolumn{1}{c|}{PODNet}  
        & 72.96 {\tiny\(\pm\)1.48} & 57.03 {\tiny\(\pm\)1.50} & 51.65 {\tiny\(\pm\)1.06} & 31.87 {\tiny\(\pm\)0.72} & 61.52 {\tiny\(\pm\)1.24} & 45.46 {\tiny\(\pm\)0.51} & 49.35 {\tiny\(\pm\)0.30} & 33.13 {\tiny\(\pm\)0.09} \\
        \multicolumn{1}{c|}{w/SAM} & 74.13 {\tiny\(\pm\)1.51} & 57.17 {\tiny\(\pm\)3.09} & 52.55 {\tiny\(\pm\)1.20} & 33.03 {\tiny\(\pm\)0.89} & 61.99 {\tiny\(\pm\)1.28} & 45.97 {\tiny\(\pm\)0.23} & 50.25 {\tiny\(\pm\)0.64} & 33.90 {\tiny\(\pm\)0.31} \\
        \multicolumn{1}{c|}{w/GAM} & 73.08 {\tiny\(\pm\)1.24} & 55.79 {\tiny\(\pm\)2.73} & 51.87 {\tiny\(\pm\)1.28} & 31.82 {\tiny\(\pm\)0.14} & 61.68 {\tiny\(\pm\)1.28} & 45.49 {\tiny\(\pm\)0.73} & 49.58 {\tiny\(\pm\)0.75} & 33.18 {\tiny\(\pm\)0.52} \\
        \multicolumn{1}{c|}{w/C-Flat} & \underline{74.19} {\tiny\(\pm\)1.22} & \underline{57.20} {\tiny\(\pm\)2.72} & \underline{53.15} {\tiny\(\pm\)1.37} & \underline{33.84} {\tiny\(\pm\)0.83} & \underline{62.21} {\tiny\(\pm\)1.14} & \underline{46.20} {\tiny\(\pm\)0.44} & \underline{50.31} {\tiny\(\pm\)0.69} & \underline{33.93} {\tiny\(\pm\)0.14} \\
        \multicolumn{1}{c|}{w/FLAD} & \textbf{74.64} {\tiny\(\pm\)1.58} & \textbf{57.62} {\tiny\(\pm\)2.55} & \textbf{53.74} {\tiny\(\pm\)1.26} & \textbf{34.85} {\tiny\(\pm\)0.42} & \textbf{62.47} {\tiny\(\pm\)1.19} & \textbf{46.92} {\tiny\(\pm\)0.66} & \textbf{50.70} {\tiny\(\pm\)0.45} & \textbf{35.04} {\tiny\(\pm\)0.39} & +0.97\%\\

    \bottomrule
    \end{tabular}
    }
    \caption{Average accuracy (\%) across all phases using 6 advanced methods, which span three categories in CL: Memory-based methods, Regularization-based methods and Expansion-based methods (w/ and w/o four optimizer plugged in). The mean and standard deviation was estimated over 3 runs. The best results are in bold and the second-best results are underlined. Average Return in the last column represents the average boost of FLAD towards C-flat in each row.}
    \label{tab:table1}
\end{table*}

\paragraph{Effect of adversarial perturbation.}
Previous work \cite{andriushchenko_towards_2022,li_friendly_2024} has demonstrated that the direction of adversarial perturbation plays a critical role in zeroth-order sharpness minimization. In particular, decomposing the perturbation into a gradient-aligned component and a stochastic-noise component has been empirically shown to improve generalization.

We extend this analysis to the first-order setting by examining how different perturbation directions affect the performance. Specifically, we consider the following variants: GAM, Pre: perturbations computed using the previous minibatch, Random: perturbation directions are randomly sampled, Full: perturbation direction aligned with the full gradient component in Eq. \ref{eq6}, Noise: perturbation direction restricted to the stochastic-noise component in Eq. \ref{eq6}, corresponding to the first-order part of FLAD. We replace the perturbation in Eq. \ref{eq9} with each component separately and evaluate their effects.

We conduct experiments using four training settings, which use ResNet-18 and Resnet-32 on CIFAR-10/100 with standard data augmentation and batchsize 128. As shown in Fig. \ref{2a}, GAM consistently improves over SGD across all settings. Importantly, Noise outperforms all other variants, indicating that the stochastic-noise component alone captures the most beneficial aspect of perturbation. Conversely, Full degrades performance, likely due to introducing conflict with learning dynamics.  

This suggests that suppressing sample-induced gradient noise improves generalization without interfering with the main optimization trajectory.
In contrast, penalizing the full gradient direction introduces conflict with learning dynamics and may degrade convergence. 
Together, these findings provide strong evidence for the effectiveness of our 
orthogonal decomposition strategy.

\paragraph{Analysis of noise components.}
To further verify the effectiveness of the stochastic-noise component, we analyze its interaction with the loss curvature using the metric \(\text{Tr}(\mathbf{H\Sigma})\), which captures the alignment between the Hessian \(\mathbf{H}\) and the gradient noise covariance \(\mathbf{\Sigma}\) \cite{zhu_anisotropic_2019}. A higher value of it suggests a stronger tendency to escape narrow minima and explore flatter regions.
We estimate \(\text{Tr}(\mathbf{H\Sigma})\) using the top Hessian eigenpairs and project update gradients along those directions. 

As shown in Fig. \ref{2b}, in early training, both the noise variants, which corresponding to the zeroth-order (FLAD-0th) and first-order (FLAD-1st) parts of FLAD, almost generally exhibits higher \(\text{Tr}(\mathbf{H\Sigma})\) than their
standard counterparts, with stronger fluctuations. This implies that the perturbation is more anisotropic and better aligned with high-curvature directions, facilitating escape from sharp minima. 
In the later training phase, the trend reverses, which indicates a reduction in curvature-aligned noise, promoting stable convergence. Consistently, the final loss landscape in Fig. \ref{2c} reveals that solutions obtained by noise variants lie in significantly flatter regions, confirming their improved generalization.
These results reinforce that the stochastic-noise component enhances generalization, offering a form of implicit regularization shared by both zeroth- and first-order sharpness minimization.

\subsection{Convergence Analysis}

We analyze the convergence properties of the our algorithm under non-convex setting in Theorem 1. 
\begin{theorem}
    Assume the loss function is twice differentiable, bounded by \(M\), and obeys the triangle inequality. Both the loss function and its second-order gradient are \(\beta\)-Lipschitz smooth. FLAD converges in all tasks with learning rate \(\eta\leq 1/\beta\), the perturbation radius \(\rho\leq 1/4\beta\), and \(\eta_i^T=\eta/\sqrt{i},\rho_i^T=\rho/\sqrt[4]{i}\) for each epoch \(i\) in any task \(T\).
\begin{equation}
    \frac{1}{n^T}\sum_{i=1}^{n^T}\mathbb{E}[\|\nabla L(w_i)\|^2]\leq \frac{C_1+\log n^T}{8\sqrt{n^T}}+\frac{64\gamma^2\sqrt{n^T}-C_2}{\beta^2n^T},
    \label{eq13}
\end{equation}
where \(n^T\) is the total iteration numbers of task \(T\), \(C_1=32M(\beta-1),C_2=32\gamma^2\) only depends on \(\gamma,M,\beta\).
\end{theorem}
See the proof in Appendix A. For non-convex stochastic optimization, Theorem 1 shows that FLAD has the convergence rate \(\mathcal{O}(\log n^T/\sqrt{n^T})\).

\begin{figure*}[htbp]
	\centering
	\begin{subfigure}{0.246\linewidth}
		\centering
		\includegraphics[width=1\linewidth]{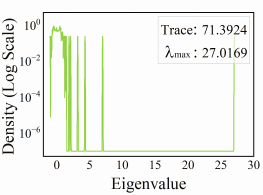}
		\caption{Replay Epoch:150 SGD}
		\label{4a}%文中引用该图片代号
	\end{subfigure}
	\centering
	\begin{subfigure}{0.246\linewidth}
		\centering
		\includegraphics[width=1\linewidth]{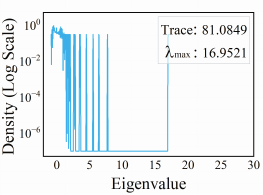}
		\caption{Replay Epoch:150 C-Flat}
		\label{4b}%文中引用该图片代号
	\end{subfigure}
	\centering
	\begin{subfigure}{0.246\linewidth}
		\centering
		\includegraphics[width=1\linewidth]{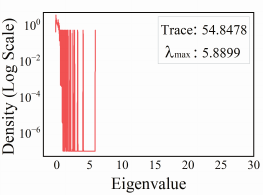}
		\caption{Replay Epoch:150 FLAD}
		\label{4c}%文中引用该图片代号
	\end{subfigure}
        \centering
	\begin{subfigure}{0.246\linewidth}
		\centering
		\includegraphics[width=1\linewidth]{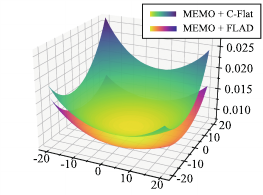}
		\caption{Loss Landscape on MEMO}
		\label{4d}%文中引用该图片代号
	\end{subfigure}
	\caption{Analysis of generalization. In (a)(b)(c), we report Hessian eigenvalue distributions and the trace of SGD, C-Flat, and FLAD on replay N=5. In (d), we show the loss landscape around final model parameters on MEMO N=10.}
	\label{fig4}
\end{figure*}

\begin{figure*}[htbp]
	\centering
	\begin{subfigure}{0.246\linewidth}
		\centering
		\includegraphics[width=1\linewidth]{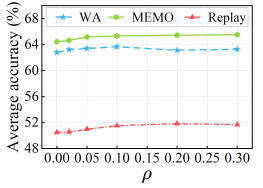}
		\caption{Ablation on \(\rho\)}
		\label{5a}%文中引用该图片代号
	\end{subfigure}
	\centering
	\begin{subfigure}{0.246\linewidth}
		\centering
		\includegraphics[width=1\linewidth]{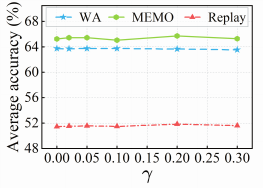}
		\caption{Ablation on \(\gamma\)}
		\label{5b}%文中引用该图片代号
	\end{subfigure}
	\centering
	\begin{subfigure}{0.246\linewidth}
		\centering
		\includegraphics[width=1\linewidth]{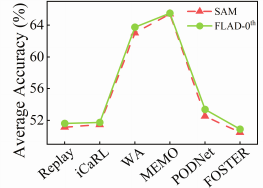}
		\caption{Ablation on FLAD-\(0^{th}\)}
		\label{5c}%文中引用该图片代号
	\end{subfigure}
        \centering
	\begin{subfigure}{0.246\linewidth}
		\centering
		\includegraphics[width=1\linewidth]{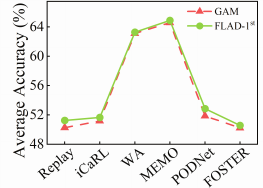}
		\caption{Ablation on FLAD-\(1^{st}\)}
		\label{5d}%文中引用该图片代号
	\end{subfigure}
	\caption{Ablation experiments. In (a)(b), we conduct ablation experiments on parameter \(\rho\) and \(\gamma\). In (c)(d), we investigate the impact of our decomposition strategy across 6 CL methods.} 
	\label{fig5}
\end{figure*}

\section{Experiments}
\subsection{Experimental Setup}

\paragraph{Datasets.} We evaluate the performance of our method on CIFAR-10, CIFAR-100~\cite{keskar_large-batch_2017} and Tiny-ImageNet~\cite{deng_imagenet_2009}. Specifically, we split CIFAR-10 into 5 tasks (2 classes per task), CIFAR-100 into 5/10 tasks (20/10 classes per task), Tiny-ImageNet into 8 tasks (25 classes per task).
We adopt two standard and widely used CL metrics: Average Accuracy (Acc), which measures the final performance by averaging classification accuracy over all \(N\) tasks at the end of training; and Average Anytime Accuracy (AAA), which averages the accuracy over all learned tasks after training on each new task.

\paragraph{Baselines.} To evaluate the efficacy of our method, we plug it into 6 leading baselines spanning the three major CL category, Replay-based methods: Replay \cite{rolnick_experience_2019} and iCaRL \cite{rebuffi_icarl_2017} ,which store raw exemplars from previous tasks, and PODNet \cite{douillard_podnet_2020} is akin to iCaRL, incorporating knowledge distillation to preserve past knowledge; Regularization-based methods: WA \cite{zhao_maintaining_2020}, which mitigates prediction bias via fairness-aware regularization; Expansion-based methods: FOSTER \cite{wang_foster_2022} and MEMO \cite{zhou_model_2023}, which dynamically expand the network using modular substructures or parameter freezing.

\paragraph{Network and training details.} All experiments are conducted on RTX 4090Ti GPUs with 96GB of RAM. For each dataset, we evaluate all methods using the same network architecture following repo \cite{zhou_pycil_2023,zhou_class-incremental_2024}. Details are provided in Appendix A. All methods are trained using vanilla SGD \cite{zinkevich_online_2003}, into which FLAD and other methods are seamlessly plugged. For fair comparison, hyperparameters \(\rho,\gamma\), \(\lambda_0\) and \(\lambda_1\) are initialized identically across tasks, while \(\lambda \in \{0.5, 0.7,0.9\}\) is tuned based on validation performance.

\begin{figure*}[htbp]
	\centering
	\begin{subfigure}{0.246\linewidth}
		\centering
		\includegraphics[width=1\linewidth]{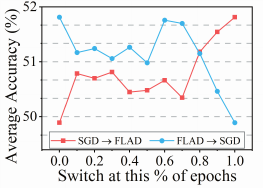}
		\caption{Replay N=10}
		\label{6a}%文中引用该图片代号
	\end{subfigure}
	\centering
	\begin{subfigure}{0.246\linewidth}
		\centering
		\includegraphics[width=1\linewidth]{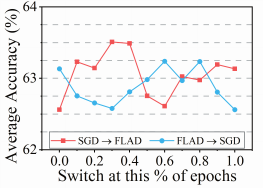}
		\caption{WA N=10}
		\label{6b}%文中引用该图片代号
	\end{subfigure}
	\centering
	\begin{subfigure}{0.246\linewidth}
		\centering
		\includegraphics[width=1\linewidth]{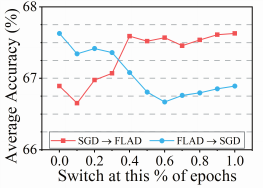}
		\caption{MEMO N=10}
		\label{6c}%文中引用该图片代号
	\end{subfigure}
        \centering
	\begin{subfigure}{0.246\linewidth}
		\centering
		\includegraphics[width=1\linewidth]{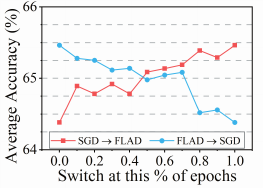}
		\caption{WA N=5}
		\label{6d}%文中引用该图片代号
	\end{subfigure}
	\caption{To further reduce overhead, we apply FLAD in a limited number of epochs within each task on different method and different settings, the horizontal axis is the transition point, before and after which we use different optimizers (SGD or FLAD).}
	\label{fig6}
\end{figure*} % , demonstrating how partial application of FLAD affects performance

\begin{figure}[t]
    \begin{subfigure}{0.487\linewidth}
    \centering
    \includegraphics[width=1\linewidth]{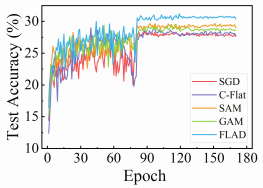}
    \caption{Convergence}
    \label{7a}%文中引用该图片代号
    \end{subfigure}
    \centering
    \begin{subfigure}{0.487\linewidth}
        \centering
        \includegraphics[width=1\linewidth]{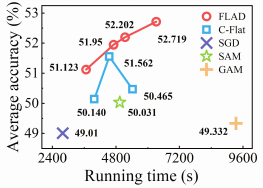}
        \caption{Training time and accuracy}
        \label{7b}%文中引用该图片代号
    \end{subfigure}
  \caption{Convergence and computation overhead. In (a), we compare test accuracy of different optimizers during training process. In (b), we compare accuracy and training time of different optimizers.}
  \label{fig7}
\end{figure}

\subsection{Performance in Continual Learning}
In this experiment, we plug our method into 6 advanced methods that cover the full range of CL methods. Table \ref{tab:table1} demonstrates that when applied across multiple benchmarks and continual settings, our method yields consistent and significant performance improvements. These enhancements manifest 1) Universality: Our method benefits all baselines, indicating its broad compatibility and minimal assumptions on the underlying CL mechanism, 2) Generalization across datasets: On each dataset, our approach consistently achieves higher final and anytime accuracies, suggesting robustness to scale and domain shift, 3) Scenario resilience: Whether under coarse- or fine-grained task splits, the gains remain stable, highlighting its adaptability across different incremental conditions. These results demonstrate that our framework serves as a versatile and effective augmentation to existing CL frameworks, offering generalization gains with negligible integration cost.

\subsection{Analysis of Generalization}

Hessian eigenvalues are widely used to quantify the sharpness of loss minima, with smaller maximum eigenvalues indicating flatter optima. 
Since flatness-aware minimization aim to reduce curvature by penalizing sharp directions, we assess whether our method indeed leads to flatter solutions by analyzing the Hessian spectrum during training.

We report the Hessian eigenvalue distributions and trace throughout CL in Fig. \ref{fig4}. Compared to vanilla SGD and C-Flat, our method yields a pronounced reduction in both the top eigenvalue and the trace, significantly suppresses high-curvature directions, indicating the discovery of flatter regions in the loss landscape. These results empirically validate that noise-aware perturbations promote flatness. 

To provide a more intuitive understanding, we visualize the loss surface by PyHessian \cite{yao_pyhessian_2020} in Fig. \ref{4d}. The landscape of MEMO becomes noticeably smoother and wider when integrated with our framework. These visual and spectral results jointly support the effectiveness of our framework in guiding the model toward flatter minima, which contributes to improved continual generalization.

\subsection{Ablation Experiments}

We conduct ablation experiments to evaluate the effect of key hyperparameters and the impact of our proposed decomposition strategy. We first investigate the sensitivity of our method to the weight of first-order term \(\gamma\) in Eq. \ref{eq14} and the perturbation radius \(\rho\) as defined in Eq. \ref{eq4} and Eq. \ref{eq9}, \(\gamma\) = 0 reduces the method to its zeroth-order variant. As shown in Fig. \ref{5a}, our method consistently outperforms the vanilla optimizer across a wide range of \(\gamma\) values, confirming the utility of incorporating first-order curvature information. Similarly, Fig. \ref{5b} illustrates the influence of \(\rho\), it demonstrate that models trained with \(\rho > 0\) uniformly outperform those without gradient ascent.

We further assess the contribution of our decomposition by comparing the performance of the decomposed and original variants across all 6 CL methods. In each case, we replace the gradient term with its decomposed variant, which isolates stochastic-noise directions. As summarized in Fig. \ref{5c} and Fig. \ref{5d}, the decomposed version consistently outperforms the original across methods. This highlights the importance of separating stochastic-noise from global trends, and affirms the value of our decomposition design in promoting generalization under continual learning.

\subsection{Convergence and Computation Overhead}

To evaluate the efficiency of our method, we analyze both convergence speed and computational cost with the Replay baseline on CIFAR-100.
As shown in Fig. \ref{7a}, our method achieves the fastest convergence among all optimizers and reaches the highest final accuracy. This demonstrates that our method not only improves generalization but also accelerates training dynamics.

To further reduce overhead, we explore applying our method only during a subset of epochs. Surprisingly, we find that even limited use of our method leads to substantial performance gains across all cases. As summarized in Fig. \ref{fig6}, applying our optimizer in just 10–20\% of epochs already yields significant improvement over vanilla SGD, and in many cases, it achieves performance comparable to or even better than using the optimizer throughout training. As a consequence, applying our method for only 30\% of the epochs reduces the computational overhead by at least 50\% compared to full-method training, highlighting its practicality and scalability for efficient continual learning. 

We compare runtime and accuracy trade-offs across different optimizers in Fig. \ref{7b}. The results show that 1) Using our method for only 10\% of the epochs (red line) already improves performance over SGD while consuming slightly less time, 2) at 20\% usage, our method outperforms GAM, SAM, and C-Flat with comparable or lower computational cost, 3) training with only 20 epochs using our method achieves higher accuracy than training with 50 epochs using C-Flat, and even 200 epochs using other optimizers.

These results confirm that our method offers a favorable balance between efficiency and effectiveness. Without introducing additional modules or increasing model complexity, it can be seamlessly integrated into existing CL methods, enabling faster convergence and better performance with fewer optimization steps.

\section{Conclusion}
In this paper, we have presented FLAD, a flatness decomposition framework for continual learning that isolates the stochastic‑noise component of sharpness‑aware perturbations. Empirical analyses reveal that this stochastic‑noise direction plays a key role in escaping sharp minima effectively and enhancing generalization. By combining zeroth‑ and first‑order decompositions with a partial application strategy, FLAD achieves strong performance gains with minimal computational overhead. Extensive experiments demonstrate that FLAD is effective across a diverse set of continual learning paradigms, highlighting its plug-and-play capability and adaptability. Its ability to improve learning efficiency with fewer optimization steps highlights the potential of curvature-guided methods for scalable and effective continual learning.

\appendix

\section{Acknowledgments}
This work was supported in part by the National Natural Science Foundation of China under Grants 62192781, 62576268, 62137002, the Key Research and Development Project in Shaanxi Province No. 2023GXLH-024, and the Project of China Knowledge Centre for Engineering Science and Technology.

% \bigskip
% \noindent Thank you for reading these instructions carefully. We look forward to receiving your electronic files!

\bibliography{Ref}

\end{document}

% \onecolumn
% \maketitle
\onecolumn
\setcounter{section}{0}
\setcounter{figure}{0}
\setcounter{table}{0}
\setcounter{theorem}{0}
\setcounter{secnumdepth}{2}
\renewcommand{\thesection}{\Alph{section}}
\renewcommand{\thesubsection}{A.\arabic{subsection}}
\renewcommand{\thefigure}{A\arabic{figure}}
\renewcommand{\thetable}{A\arabic{table}}
% \maketitle
\section*{\LARGE{Beyond Sharpness: A Flatness Decomposition Framework for Efficient Continual Learning}}
\vspace{0.5em}
\section*{Supplementary Material}
\vspace{0.5em}

\subsection{Overview}

% In this supplementary material, we first present the proof of convergence (Appendix A.2). Next, we present some additional experiments including effect of batchsize on first-order sharpness minimization and more visualizations at (Appendix A.3). And finally, we provide more details on the experiments (Appendix A.4)
In this supplementary material, we first present the proof of convergence (Appendix A.2). Next, we provide additional experiments, including the effect of batchsize on first-order sharpness minimization, as well as more visualizations from the final stages and throughout the training process (Appendix A.3). Finally, we offer further details on our experimental setup (Appendix A.4).

\subsection{Proof of convergence}
\textit{proof.}
\begin{assumption}
    The loss function is twice differentiable, bounded by \(M\), and obeys the triangle inequality. Both the loss function and its second-order gradient are \(\beta\)-Lipschitz smooth. Learning rate \(\eta\leq 1/\beta\), the perturbation radius \(\rho\leq 1/4\beta\), and \(\eta_i^T=\eta/\sqrt{i},\rho_i^T=\rho/\sqrt[4]{i}\) for each epoch \(i\) in any task \(T\).
\end{assumption}

% \frac{2\beta M\eta_0+\rho_0^2\beta^3\eta_0+\rho_0^2\beta^2\log n}{\sqrt{n}}with bounded variance \(\sigma^2\), 
With Assumptions 1, the convergency of zeroth-sharpness is guaranteed \cite{li_friendly_2024} by
\begin{align}
    \frac{1}{n}\sum_{i=1}^{n}\mathbb{E}[\|\nabla L^{R_\rho^0}(w_i)\|^2]&\leq\frac{2\beta}{\sqrt{n}}\mathbb{E}[L(w_0)-L(w^*)]+\frac{32M+1}{16\sqrt{n}}\sum_{i=1}^{n}\mathbb{E}[i^{-\frac{1}{2}}]+\frac{1}{16\sqrt{n}}\sum_{i=1}^{n}\mathbb{E}[i^{-1}]\\
    &\leq\frac{2\beta}{\sqrt{n}}\mathbb{E}[L(w_0)-L(w^*)]+\frac{\log n-32M}{16\sqrt{n}}
\end{align}
thus, the zeroth-order part of FLAD is bounded:
\begin{align}
    \frac{1}{n^T}\sum_{i=1}^{n^T}\mathbb{E}[\|\nabla L^{R_\rho^0}(w_i)\|^2]&\leq\frac{2\beta}{\sqrt{n^T}}[L(w^T)]+\frac{\log n-32M}{16\sqrt{n}}\\
    &\leq\frac{2\beta M}{\sqrt{n^T}}+\frac{\log n^T-32M}{16\sqrt{n^T}}
\end{align}

\begin{theorem}
    let \(\delta=\rho_i\frac{\nabla\|\nabla L(w)\|-n_i}{\|\nabla\|\nabla L(w)\|-n_i\|}\), with Assumptions 1, the first-order part is bounded by
    \begin{align}
        &\qquad\frac{1}{n^T}\sum_{i=1}^{n^T}\mathbb{E}[\|\nabla L^{R_\rho^1}(w_i)\|^2]\leq\frac{1}{n^T}\sum_{i=1}^{n^T}\mathbb{E}[\|\nabla L(w^T+\delta)-\nabla L(w^T)\|^2] \\
    &\leq\frac{\beta^2}{n^T}\sum_{i=1}^{n^T}\mathbb{E}[\|\delta\|^2]
    \leq\frac{\beta^2\eta^2}{n^T}\sum_{i=1}^{n^T}\mathbb{E}[\|\rho_i^T\|^2]
    \leq\frac{\rho^2}{n^T}\sum_{i=1}^{n^T}\mathbb{E}[i^{-\frac{1}{2}}]
    \leq\frac{16(2\sqrt{n^T}-1))}{\beta^2n^T}
    \end{align}
\end{theorem}

\begin{theorem}
    with Assumptions 1, by combining the zeroth- and first-order parts, we can prove FLAD converges in all tasks that
    \begin{align}
        \frac{1}{n^T}\sum_{i=1}^{n^T}\mathbb{E}[\|\nabla L(w_i)\|^2]&\leq\frac{2}{n^T}\sum_{i=1}^{n^T}\mathbb{E}[\|\nabla L^{R_\rho^0}(w_i)\|^2]+\frac{2}{n^T}\sum_{i=1}^{n^T}\mathbb{E}[\|\gamma\nabla L^{R_\rho^1}(w_i)\|^2] \\
        &\leq\frac{4\beta M}{\sqrt{n^T}}+\frac{32\gamma^2(2\sqrt{n^T}-1))}{\beta^2n^T}+\frac{\log n^T-32M}{8\sqrt{n^T}} \\
        &\leq\frac{\log n^T+32M(\beta-1)}{8\sqrt{n^T}}+\frac{32\gamma^2(2\sqrt{n^T}-1))}{\beta^2n^T}
    \end{align}
\end{theorem}

\begin{figure}[h]
  \centering
  \includegraphics[width=0.3\linewidth]{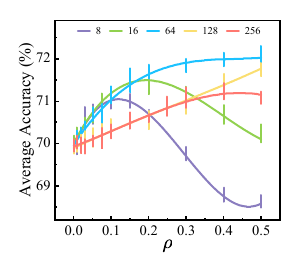}
  \caption{The average accuracy of GAM with a range of batchsizes.}
  \label{app1}
\end{figure}

\subsection{Additional experiments}
\paragraph{Effect of batchsize.}
% We verify the sensitivity to batchsize which suggests combing zeroth- and first-order sharpness.
We analyze the sensitivity to batchsize, which provides motivation for combining zeroth- and first-order sharpness.
In practice, sharpness is compute over a subset of data points, typically corresponding to batchsize during training.
While prior work on zeroth-order flatness \cite{foret_sharpness-aware_2021} has demonstrated that smaller batchsizes lead to flatter minima and better generalization.
To investigate the behavior of first-order sharpness, we train a small model using first-order sharpness minimization with a range of values of batchsize. As shown in Fig. \ref{app1}, smaller batch sizes consistently improve generalization, resembling the trend observed in zeroth-order flatness. 
However, excessively small batches often lead to inefficient hardware utilization. In practice, a moderate batch size (e.g., 128) provides a favorable trade-off between generalization and efficiency, supporting the practical benefits of combining zeroth- and first-order sharpness.

\paragraph{More visualizations.}We present additional visualizations of the loss landscape in two cases using PyHessian \cite{yao_pyhessian_2020}: 1) changes in the loss landscape at the final stage of training, and 2) changes across tasks during continual learning.

\begin{figure*}[h]
	\centering
	\begin{subfigure}{0.325\linewidth}
		\centering
		\includegraphics[width=1\linewidth]{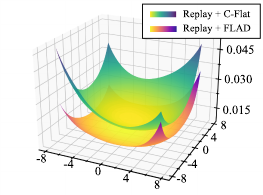}
		\caption{Replay}
		\label{a2a}%文中引用该图片代号
	\end{subfigure}
	\centering
	\begin{subfigure}{0.325\linewidth}
		\centering
		\includegraphics[width=1\linewidth]{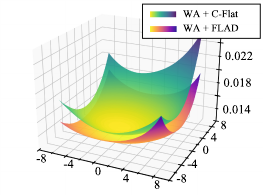}
		\caption{WA}
		\label{a2b}%文中引用该图片代号
	\end{subfigure}
    \centering
	\begin{subfigure}{0.325\linewidth}
		\centering
		\includegraphics[width=1\linewidth]{figure/3dnew.pdf}
		\caption{MEMO}
		\label{a2c}%文中引用该图片代号
	\end{subfigure}
	\caption{The visualizations of loss landscapes at the final stage of training.}
	\label{app2}
\end{figure*}

We first focus on a small neighborhood around the final model parameters to capture fine-grained differences in the loss landscape. We visualize three representative methods including Replay \cite{rolnick_experience_2019}, WA \cite{zhao_maintaining_2020}, and MEMO \cite{zhou_model_2023}, which is selected from different categories of CL approaches. As shown in Fig.~\ref{app2}, in this detailed view, FLAD consistently leads to a flatter loss landscape compared to the state-of-art method C-Flat, more intuitively revealing its underlying mechanism.

\begin{figure*}[h]
	\centering
	\begin{subfigure}{0.325\linewidth}
		\centering
		\includegraphics[width=1\linewidth]{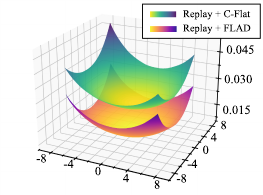}
		\caption{Task1}
		\label{a2a}%文中引用该图片代号
	\end{subfigure}
	\centering
	\begin{subfigure}{0.325\linewidth}
		\centering
		\includegraphics[width=1\linewidth]{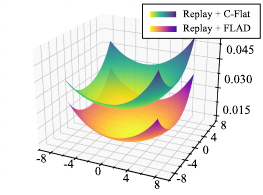}
		\caption{Task2}
		\label{a2b}%文中引用该图片代号
	\end{subfigure}
    \centering
	\begin{subfigure}{0.325\linewidth}
		\centering
		\includegraphics[width=1\linewidth]{figure/replay.pdf}
		\caption{Task4}
		\label{a2c}%文中引用该图片代号
	\end{subfigure}
	\caption{The visualizations of loss landscapes during continual learning.}
	\label{app3}
\end{figure*}

Second, we further visualize the loss landscape across multiple tasks to provide more intuitive insights into the training dynamics during continual learning. For simplicity, we select Replay \cite{rolnick_experience_2019} and set \(N=5\). We visualize the loss surface on Tasks 1, 2, and 4. As shown in Fig.~\ref{app3}, the loss landscapes become significantly flatter compared to C-Flat across all tasks. This consistent trend provides strong empirical evidence supporting the effectiveness of FLAD.

\subsection{Implementation detail}

In all experiments, we train deep networks using SGD with a step size of \(0.1\), momentum of \(0.9\), and \(\ell_2\)-regularization parameter \(0.0005\). For other optimizers, we use the same step size and momentum, but set \(\ell_2\)-regularization parameter \(0.0002\). All datasets are augmented using standard techniques: random cropping and horizontal flipping. We use a batch size of \(128\) in most experiments unless otherwise specified.

For architecture, we adopt a pre-activation ResNet-18 for Tiny-ImageNet and a ResNet-32 for CIFAR-10/100, both with a width factor of 64. Learning rates follow a piecewise constant schedule. For models trained with SGD, the learning rate decays by a factor of 10 at 30\%, 60\%, and 85\% of the total training epochs. For other optimizers, the first decay typically occurs at around 45\% of the training duration. Models trained with SGD are trained for 200 epochs, while those using other optimizers are trained for 170 epochs.

For all experiments involving sharpness-aware methods, we set the perturbation radius \(\rho=0.2\), except for FOSTER in the continual learning experiments, where \(\rho=0.02\) is used. All experiments are conducted on a single GPU. To ensure robustness, many results are replicated across different random seeds.

\bibliography{Appref}